\documentclass[11pt]{article}
\usepackage{spconf,amsmath,graphicx}
\usepackage{booktabs}
\usepackage{multirow}
\usepackage{hyperref}
\usepackage{INTERSPEECH_mod}

\title{Spontaneous Informal Speech Dataset for Punctuation Restoration}
\name{Xing Yi Liu$^1$, Homayoon Beigi$^{2,3}$}
\address{$^1$University of Waterloo, Ontario, Canada\\ 
         $^2$Recognition Technologies, Inc., New York, USA\\
         $^3$Columbia University, New York, USA\\ \ \\
  \small{Technical Report: \href{https://www.recognitiontechnologies.com/techreport/RTI-20240917-01.pdf}{RTI-20240917-01}}\\
    \small{\href{http://dx.doi.org/10.13140/RG.2.2.33136.88329}{DOI: 10.13140/RG.2.2.33136.88329}}
}
\email{$^1$xy39liu@uwaterloo.ca, $^2$beigi@recotechnologies.com}
\begin{document}
%
\maketitle
\begin{abstract}
Presently, punctuation restoration models are evaluated almost solely on well-structured, scripted corpora. On the other hand, real-world ASR systems and post-processing pipelines typically apply towards spontaneous speech with significant irregularities, stutters, and deviations from perfect grammar. To address this discrepancy, we introduce SponSpeech, a punctuation restoration dataset derived from informal speech sources, which includes punctuation and casing information. In addition to publicly releasing the dataset, we contribute a filtering pipeline that can be used to generate more data. Our filtering pipeline examines the quality of both speech audio and transcription text. We also carefully construct a ``challenging" test set, aimed at evaluating models' ability to leverage audio information to predict otherwise grammatically ambiguous punctuation. SponSpeech is available at \url{https://github.com/GitHubAccountAnonymous/PR}, along with all code for dataset building and model runs.
\end{abstract}
\begin{keywords}
Punctuation restoration, speech recognition post-processing, dataset, corpus
\end{keywords}

\section{Introduction}
\label{sec:introduction}

Researchers in the field of punctuation restoration widely recognize the task's value: By supplying punctuation information to automatic speech recognition (ASR) systems' raw, unpunctuated output, we aid execution of downstream tasks like machine translation, natural language understanding, and much more. The field has made great progress in recent history, from proposals of recurrent neural networks \cite{tilk2016bidirectional} and end-to-end approaches \cite{nozaki2022end} to adaptations of transformers \cite{zhu2022unified} and time-delay neural networks \cite{liu2024efficient}. However, an emerging problem is the singular type of datasets used for evaluating English punctuation restoration models.

In the English language domain, research on text-only punctuation models \cite{lai2023boosting, huang2021token, alam2020punctuation, che2016punctuation} almost always use the IWSLT 2011 and 2012 datasets, both derived from TED talks' transcriptions. In the other category, recent models considering acoustics and text \cite{nozaki2022end, zhu2022unified, liu2024efficient, kim2023improved} has very often used the MuST-C dataset \cite{di2019must}, also derived from TED talks' audio and transcriptions. Each of these models presented steady improvements in performance and indeed advanced the field. The dataset on which their insights are based, however, is limited in scope. TED talks are scripted monologues that are carefully practiced by the presenters before formal delivery. Samples in the IWSLT and MuST-C datasets therefore exhibit high uniformity in style. Audio and transcriptions derived from this source have high quality but lack many natural traits of typical speech. For example, stutters, hesitations, and other spoken imperfections should optimally present themselves often in punctuation restoration data in order for trained models to robustly handle them. Although some conversations between the TED host and presenters are available in small numbers, the IWSLT and MuST-C datasets predominantly feature flawless, formal monologues.

To address this limitation in data for training and evaluating punctuation restoration models, we create the SponSpeech dataset. We focus on extracting spontaneous, informal speech, which more accurately reflects natural conversations, rather than formal speeches that rarely occur in daily life. We take podcasts as our dataset source, since they offer stylistic diversity. They contain not only stints of discussion on focused subjects, but also interactive dialogue between multiple speakers. Importantly, spoken imperfections are preserved in podcast conversations.

In addition to releasing SponSpeech as a punctuation restoration dataset of varied style, our contribution has another intention: to emphasize utterances with punctuation ambiguity. A significant value of performing punctuation restoration is to improve readability of the text \cite{tundik2018user}. Partially, this involves eliminating ambiguity. For example, the text
\begin{center}
    \textit{i have eight boys}
\end{center}
has \textbf{punctuation ambiguity}, since ``\textit{I have eight boys.}" suggests that the speaker has eight sons, whereas ``\textit{I have eight, boys.}" suggests that the speaker is indicating to their boys that they possess eight objects. On the other hand, punctuation restoration has much less value in a text like
\begin{center}
    \textit{i have an apple}
\end{center}
as the only way to apply punctuation is: ``\textit{I have an apple.}" There is no punctuation ambiguity in this latter case.

In building our dataset, we aim to include a greater number of utterances with punctuation ambiguity. In particular, we create two test sets, one with slightly more ambiguous cases than the other. Such a dataset will better evaluate models' ability to leverage audio patterns to resolve punctuation ambiguities. As a simple example, a longer pause in certain contexts may distinguish the absence or presence of a comma.

\section{Related Works}
\label{sec:related}

To train punctuation restoration models, only corpora with punctuation information are useful. This is particularly noteworthy, because many ASR datasets' transcriptions are all lower case with no punctuation, and hence not suitable for punctuation restoration. In the case of text-only models, most written text are acceptable, but this paper focuses on the domain that considers both text and acoustics. Prior work has convincingly confirmed the added value of incorporating acoustics information \cite{yi2019self}, and certainly ASR systems can be inexpensively streamlined to output audio embeddings for punctuation purposes \cite{liu2024efficient}.

The IWSLT pure-text dataset has approximately 2.4 million words, which corresponds to a rough estimate of 267 hours of ``average" speech \cite{brigance1926fast}. For some of the most recognized datasets with speech audio and that have punctuation information, see the summary provided in Table \ref{tab:datasets}. 

\begin{table}[t]
    \caption{Summary of speech datasets suitable for punctuation restoration. Fisher corpus refers to the punctuated section used by \cite{sunkara2020multimodal}.}
    \label{tab:datasets}
    \centering
    \begin{tabular}{ c c c }
        \toprule
        \textbf{Dataset} & \textbf{Hours} & \textbf{Style} \\
        \midrule
        SponSpeech & 665 & Podcasts \\
        Libriheavy \cite{kang2024libriheavy} & 56389 & Reading \\
        NSC \cite{koh2019building} & 2170 & Reading \\
        LibriTTS \cite{zen2019libritts} & 585 & Reading \\
        MuST-C v1 \cite{di2019must} & 438 & Monologues \\
        Fisher \cite{cieri2004fisher} & 432 & Telephone \\
        Switchboard-1 \cite{godfrey1992switchboard} & 260 & Telephone \\
        \bottomrule
    \end{tabular}
\end{table}

The largest dataset, Libriheavy, impressively contains over 50000 hours of audio and punctuation/casing information. A dataset of book recordings, its speech content has a scripted and uniformly flawless style. Similarly, NSC and LibriTTS involve readers verbalizing prior written texts. As discussed, MuST-C comprises formal monologues in the form of TED talks.

Among the most popularly used datasets, only the Fisher and Switchboard-1 corpora feature spontaneous speech. Unfortunately, they require a paid membership to the Linguistic Data Consortium or a non-member licensing fee to access. We release a reasonably abundant 665 hours of speech data with punctuated transcripts under a Creative Commons Attribution-NonCommercial 4.0 International (CC BY-NC 4.0) license, allowing all researchers to make use of our resource for free.

\section{SponSpeech}
\label{sec:sponspeech}

The SponSpeech dataset is released with four standard subsets in order to facilitate reproducibility of future research: train, dev (validation), test, and test-amb. The last subset, test-amb, is an evaluation set with slightly more cases of punctuation ambiguity. We expect models to perform slightly worse on test-amb than on the ordinary test set, as resolving ambiguities presents an extra challenge. Special insights must be drawn from acoustic information when, grammatically, multiple versions of punctuation are possible. At the same time, we do not desire for test-amb to drastically differ from utterance population norms. A dataset and its subsets should accurately reflect the state of the entire represented population \cite{ribeiro2016should}, a fact we strived to balance with creating a challenging, second evaluation set.

Table \ref{tab:subsets} lists statistics on each subset. We source our data from YouTube videos published under Creative Commons Attribution 4.0 International Licenses, rather than standard YouTube licenses. This respects all uploaders' rights, while allowing our creation of SponSpeech as an adapted work with appropriate attribution. The fourth column reports the number of videos involved in each subset. Videos are sliced to create utterances.

The train, dev, test, and test-amb sets make up approximately 70\%, 12\%, 9\%, and 9\% of the entire dataset, respectively. The overall minimum, average, and maximum utterance durations are 1.6 s, 11.6 s, and 45.0 s, respectively.

\begin{table}[t]
    \caption{Statistics on each subset of SponSpeech.}
    \label{tab:subsets}
    \centering
    \begin{tabular}{ c c c c }
        \toprule
        \textbf{Subset} & \textbf{Hours} & \textbf{Utterances} & \textbf{Videos} \\
        \midrule
        train & 469 & 147209 & 1736 \\
        dev & 79 & 25253 & 277 \\
        test & 58 & 17697 & 205 \\
        test-amb & 60 & 16473 & 397 \\
        \bottomrule
    \end{tabular}
\end{table}

The following subsections describe the dataset creation process.

\subsection{Data Source}
\label{ssec:source}

Each YouTube video has a unique video ID. To obtain an initial pool of YouTube video IDs, we specify a filter code in the search query URL (in this case, \texttt{sp=EgQoATAB}) that only allows videos with a Creative Commons license and subtitles to be returned. The resulting list of videos merely serves as candidates for inclusion in SponSpeech, with careful evaluation still needed for each video to ensure criteria for desirable properties are met. Then, a series of five filters are applied to eliminate unacceptable candidates.

We use the \texttt{yt-dlp} tool to download videos' subtitle and audio content, as well as metadata.

\begin{figure*}[t] 
    \centering 
    \includegraphics[width=17cm]{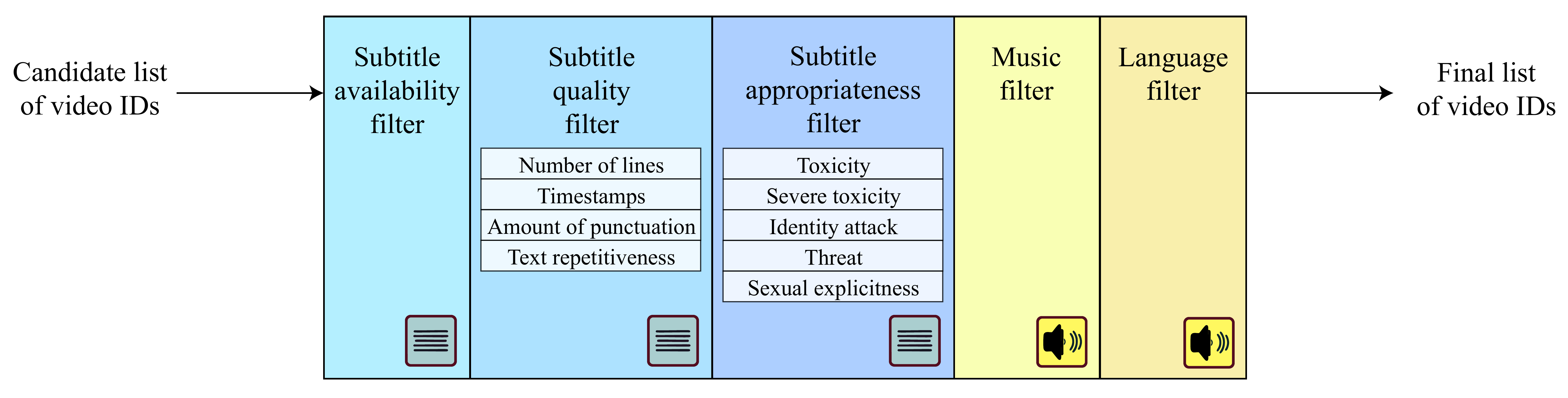}
    \caption{Filtering pipeline used to create SponSpeech. Blue indicates text-based filter, and yellow indicates audio-based filter, also shown by the bottom-right icons. Sub-boxes within the subtitle quality and appropriateness filters are the evaluation criteria used.}
    \label{fig:filters}
\end{figure*}

\subsection{Filters}
\label{ssec:filters}

Figure \ref{fig:filters} provides an illustration of the filtering pipeline. Each filter acts on the pool of candidate video IDs. Since text can be more computationally efficient to analyze than audio, the first three filters use text as the basis of evaluation. The fourth and fifth filters evaluate audio.

First, the \textbf{subtitle availability filter} simply detects whether manually-uploaded English transcriptions exist for each candidate video. YouTube's system automatically generates subtitles, but we do not wish for ASR output text to be used as ground-truth text labels in SponSpeech. Instead, this filter requires the content creator to have written and supplied the video's transcription themself.

Second, the \textbf{subtitle quality filter} assesses four criteria evaluating the content of subtitles. Subtitle files follow a standard format, with a header block of metadata to begin, then roughly alternating lines of timestamp ranges and the corresponding text. As a side note, subtitle files provide, by default, accurate alignments to serve as the basis for later extraction of individual utterances. These timestamps are either provided manually by the uploader alongside the transcription, or by YouTube's alignment algorithm, depending on the uploader's choice.
\begin{enumerate}
    \item \textit{Number of lines.} Each video's subtitle file is required to have at least 20 lines, a heuristic selected on the basis of manual examination and erring on the side of caution. Each subtitle file has a header block with metadata. As such, videos with substantial content most often have, at the very minimum, 20 subtitle file lines.
    \item \textit{Timestamp formatting.} Each video's subtitle file must contain timestamps that follow a standard format common to the vast majority of videos. Unfortunately, a small number have highly irregular formatting, such as those with mistakenly littered HTML tags. These are difficult to consistently interpret and hence filtered out.
    \item \textit{Amount of punctuation.} Some videos' subtitle files are poorly written and unexplicably have very few punctuation marks. This presents a problem for our dataset, which is designed for punctuation restoration. As a result, this criteria checks for a minimum of one punctuation mark for every ten subtitle lines. 
    \item \textit{Text repetitiveness.} Some videos' subtitle files have much overlapping text between different lines. In other words, the same phrase may appear repetitively in error, with the corresponding audio not featuring any repetition. Each line has a distinct corresponding timestamp range; the same text appearing on multiple lines may indicate alignment issues. Videos with text repetitiveness of this type (and not real repetitiveness) are filtered out.
\end{enumerate}

Third, the \textbf{subtitle appropriateness filter} eliminates videos with inappropriate language. The types of inappropriate language targeted include toxicity, severe toxicity, identity attack, threat, and sexual explicitness. While these samples do exist in the natural population of speech data, we do not wish for such language to be further propagated through a public dataset. We pass each video's subtitles through the \texttt{Detoxify} model in Python's \texttt{detoxify} library \cite{hanu2020detoxify, hanu2021how}. With probability outputs describing each inappropriateness category, we eliminate videos exceeding a threshold.

Fourth, the \textbf{music filter} ensures that only speech-type audio content is accepted into the dataset. Even though the search keyword ``podcast" was applied to obtain candidate videos, YouTube contains both speech-dominant and music-dominant content. We use a wav2vec 2.0-based speech/music classifier to determine the amount of speech, music, and simultaneous content in each video \cite{baevski2020wav2vec, demirkıran2023wav2vec2}. Those with more music than speech, measured based on a probability heuristic, are filtered out. Note that not all videos with music present should be eliminated, because, for example, speech overlayed on top of quieter music has significant validity as an utterance to train a punctuation restoration model.

Fifth and last, the \textbf{language filter} allows only English content to pass through, for the scope of our intended dataset. A Whisper-based language identification model evaluates each video's speech audio \cite{radford2023robust, gandhi2023whisper}.

We deemed it acceptable to involve community-trained machine learning models, albeit ones with convincing test set results and fine-tuned from established foundation models, because any performance drawbacks could be compensated by using highly cautious threshold values for filtering decisions. We always select threshold values (including in sub-filter criteria) conservatively to favor eliminating unproblematic videos, rather than failing to eliminate problematic ones. In other words, the aim is to maximize recall and somewhat overlook precision, in the context of a positive class being videos for elimination. The rationale is that we can easily source more candidate videos, but we must be careful not to allow unwanted samples to enter the dataset.

Videos that pass all filters then form utterances in SponSpeech.

\subsection{Creating Utterances}
\label{ssec:utterances}

Utterances are the unit basis of data samples in SponSpeech. For each video that passed through all filters, select sentence delimiters are used to split the video into utterances. As it has become standard in punctuation restoration research to consider only the dominant punctuation marks full stop (.), comma (,), and question mark (?), we use full stops and question marks as split points.

Utterances naturally end in terminal punctuation marks, a standard used to create nearly all ASR datasets \cite{kang2024libriheavy}. However, full stops and question marks are also made available in the middle of utterances, so that punctuation restoration models can be trained robustly on SponSpeech.

The result of this step is a pool of utterances with audio extracted using YouTube-provided alignment information. Concretely, utterances are \texttt{.wav} files with single-channel audio of 16 kHz sample rate. The pool is split into the four data subsets: train, dev, test, and test-amb. As previously mentioned, test-amb is a special test set with an increased number of utterances with punctuation ambiguity. The following subsection describes the creation of test-amb.

\subsection{Creating Ambiguous Test Set}
\label{ssec:ambiguous}

To create test-amb, the main consideration is determining whether each utterance's text contains punctuation ambiguity, effectively forming a binary classification task. The positive class is ``has punctuation ambiguity", and the negative class is ``has no punctuation ambiguity." For the meaning and an example of punctuation ambiguity, please refer to Section \ref{sec:introduction}.

The framework with which we determine punctuation ambiguity is ELECTRA \cite{clark2020electra}, fine-tuned on a small number of samples. To accomplish this, we manually label 100 samples for training and 20 samples for validation, with a few examples shown in Table \ref{tab:electra}. Each sample's input is unpunctuated utterance text, and we call ``label A" the list of all correct versions of punctuation, separated by newline characters. ``Label B" is binary, with 0 being ``has no punctuation ambiguity" (i.e. only one item in label A) and 1 being ``has punctuation ambiguity" (i.e. more than one item in label A).

\begin{table}[t]
    \caption{Data samples used to fine-tune ELECTRA for creating test-amb. Label A is manually created and lists all correct versions of punctuation for the input. Label B is binary and indicates whether there is punctuation ambiguity.}
    \label{tab:electra}
    \centering
    \begin{tabular}{ c c c }
        \toprule
        \textbf{Input} & \textbf{Label A} & \textbf{Label B} \\
        \midrule
        and i liked the fact... & And I liked the fact... & 0 \\
        \midrule
        \multirow{2}{*}{... go there but you...} & ... go there, but you... & \multirow{2}{*}{1} \\
        & ... go there. But you... & \\
        \midrule
        \multirow{2}{*}{... then education is...} & ... then education is... & \multirow{2}{*}{1} \\
        & ... then, education is... & \\
        \bottomrule
    \end{tabular}
\end{table}

ELECTRA utilizes a generator-discriminator architecture. We leverage both components for binary classification as follows:
\begin{enumerate}
    \item \textbf{Generator tuning for text generation.} The input, unpunctuated utterance text and label A are used to fine-tune ELECTRA's generator for text generation. The model is prompted to generate all correct versions of punctuation based on the given text. This step's task is more complex than ultimately needed, since only a final binary prediction is used to create test-amb. However, this precursor step is designed to guide the foundation model towards reasoning about punctuation. Directly predicting label B without this step may obfuscate whether the task relates to punctuation at all.
    \item \textbf{Generator for ambiguity classification.} After performing step 1, the generator is used to obtain last hidden layer embeddings for the input text. Along with label B, these embeddings are used to train a multilayer perceptron (MLP) for binary classification. The results of this step -- predictions about whether utterances contain punctuation ambiguity -- are directly used in the creation of test-amb.
    \item \textbf{Discriminator for ambiguity classification.} Similar to step 2, the discriminator's last hidden layer embeddings for the input text and label B are used to train a MLP for binary classification. Since the discriminator is not designed for text generation, its usage does not involve the precursor fine-tuning step as in the generator.
\end{enumerate}

\begin{table*}[th]
  \caption{Precision (P), recall (R), and F1 score (F1) achieved by models trained on MuST-C and tested on SponSpeech.}
  \label{tab:cross-sponspeech}
  \centering
  \begin{tabular}{ c | c | c c c | c c c | c c c | c c c }
    \toprule
    \multirow{2}{*}{\textbf{Subset}} & \multirow{2}{*}{\textbf{Model}} & \multicolumn{3}{c|}{\textbf{Full Stop}} & \multicolumn{3}{c|}{\textbf{Comma}} & \multicolumn{3}{c|}{\textbf{Question}} & \multicolumn{3}{c}{\textbf{Overall}} \\
    & & P & R & F1 & P & R & F1 & P & R & F1 & P & R & F1 \\
    \midrule
    \multirow{2}{*}{test} & EfficientPunct & 76.9 & 84.4 & 80.5 & 54.6 & 76.1 & 63.6 & 70.5 & 79.7 & 74.8 & 63.2 & 79.7 & 70.5 \\
    & BERT & 75.5 & 85.3 & 80.1 & 60.7 & 69.3 & 64.7 & 69.4 & 80.0 & 74.3 & 67.1 & 76.3 & 71.4 \\
    \midrule
    \multirow{2}{*}{test-amb} & EfficientPunct & 76.6 & 80.6 & 78.6 & 55.1 & 76.2 & 63.9 & 67.4 & 77.7 & 72.2 & 63.1 & 78.1 & 69.8 \\
    & BERT & 75.9 & 81.7 & 78.7 & 61.0 & 69.6 & 65.0 & 66.9 & 79.2 & 72.6 & 67.2 & 75.0 & 70.9 \\
    \bottomrule
  \end{tabular}
\end{table*}

\begin{table*}[t]
  \caption{Precision (P), recall (R), and F1 score (F1) achieved by models trained on SponSpeech and tested on MuST-C. The test set used is the same as in \cite{zhu2022unified, liu2024efficient}.}
  \label{tab:cross-mustc}
  \centering
  \begin{tabular}{ c | c c c | c c c | c c c | c c c }
    \toprule
    \multirow{2}{*}{\textbf{Model}} & \multicolumn{3}{c|}{\textbf{Full Stop}} & \multicolumn{3}{c|}{\textbf{Comma}} & \multicolumn{3}{c|}{\textbf{Question}} & \multicolumn{3}{c}{\textbf{Overall}} \\
    & P & R & F1 & P & R & F1 & P & R & F1 & P & R & F1 \\
    \midrule
    EfficientPunct & 79.3 & 81.8 & 80.5 & 64.4 & 76.7 & 70.0 & 82.9 & 77.6 & 80.1 & 70.7 & 78.9 & 74.5 \\
    BERT & 82.0 & 76.7 & 79.3 & 69.1 & 65.7 & 67.3 & 85.3 & 73.4 & 78.9 & 75.0 & 70.5 & 72.7 \\
    \bottomrule
  \end{tabular}
\end{table*}

Based on optimal performance on the validation set, a weight of $\alpha = 0.8$ is assigned to the generator's sigmoid output probabilities, and a weight of $1-\alpha = 0.2$ is assigned to the discriminator's. As expected, the generator, tuned with the precursor generation step, is more useful. The weighted sum of probability values are used for the final predictions of punctuation ambiguity on all SponSpeech utterances. A threshold of $\beta = 0.6$ was similarly selected based on validation, with prediction probabilities $p \geq \beta$ being classified as ``having punctuation ambiguity."

A formal evaluation of the fine-tuned models were not deemed necessary, since the generator, discriminator, and MLPs only needed to give a general suggestion as to which utterances likely had punctuation ambiguity. Maximizing accuracy was less important, unlike with typical machine learning tasks, since classification mistakes do not deter the overall formation of test-amb. Moreover, any additional samples labeled for a potential test set would much better serve to enlarge the scant fine-tuning and validation data subsets anyway.

In the end, test-amb is created with a greater number of utterances predicted to have punctuation ambiguity. The effectiveness of this creation procedure is proven in the following sections' results.

\section{Evaluation}

To evaluate the quality of SponSpeech, we follow a procedure similar to that used for MuST-C v2 \cite{cattoni2021must}. An indication of dataset quality is achieving reasonable results when cross-testing with another dataset, i.e. (1) training on another dataset and testing on SponSpeech, and (2) training on SponSpeech and testing on another dataset. We use MuST-C for this purpose, since it is currently the most frequently used multimodal (with text and acoustics) dataset for punctuation restoration. The models picked are BERT \cite{devlin2018bert} and EfficientPunct \cite{liu2024efficient}, which are among the recent top-performing models in the text-only and multimodal domains.

As shown in Table \ref{tab:cross-sponspeech}, both models trained on MuST-C achieve overall F1 scores around or above 70\% when testing on SponSpeech. By comparison against within-SponSpeech dataset results in Tables \ref{tab:results-test} and \ref{tab:results-test-amb}, MuST-C-trained and SponSpeech tested results appear reasonable. Namely, not training on the SponSpeech population itself only decreases overall F1 score on SponSpeech's test and test-amb sets by about 2.3\% to 4.0\%. The quality of SponSpeech's test and test-amb sets are hence signaled. As expected, results for test-amb are lower than for the test set, demonstrating the effectiveness of our ELECTRA tuning process in creating a test set with more punctuation ambiguity.

For results of training on SponSpeech and testing on MuST-C shown in Table \ref{tab:cross-mustc}, overall F1 scores achieved are well above 70\%, demonstrating the quality of SponSpeech's training and validation sets. Together, the two sets of cross-testing results show that SponSpeech has high quality and robustness.

\section{Baselines}

\begin{table*}[t]
  \caption{Precision (P), recall (R), and F1 score (F1) achieved by models on SponSpeech's test set.}
  \label{tab:results-test}
  \centering
  \begin{tabular}{ c | c c c | c c c | c c c | c c c }
    \toprule
    \multirow{2}{*}{\textbf{Model}} & \multicolumn{3}{c|}{\textbf{Full Stop}} & \multicolumn{3}{c|}{\textbf{Comma}} & \multicolumn{3}{c|}{\textbf{Question}} & \multicolumn{3}{c}{\textbf{Overall}} \\
    & P & R & F1 & P & R & F1 & P & R & F1 & P & R & F1 \\
    \midrule
    EfficientPunct & 79.3 & 85.6 & 82.3 & 56.9 & 82.6 & 67.4 & 74.0 & 80.4 & \textbf{77.0} & 65.2 & 83.8 & 73.3 \\
    UniPunc & 70.5 & 83.2 & 76.3 & 46.9 & 75.9 & 58.0 & 71.8 & 71.2 & 71.5 & 55.8 & 78.7 & 65.3 \\
    BERT & 82.4 & 82.3 & \textbf{82.4} & 66.5 & 75.0 & \textbf{70.5} & 76.1 & 77.6 & 76.8 & 73.0 & 78.1 & \textbf{75.4} \\
    GPT-4 Turbo & 87.0 & 70.0 & 77.5 & 47.7 & 85.0 & 61.1 & 67.5 & 81.9 & 74.0 & 58.0 & 78.7 & 66.8 \\
    \bottomrule
  \end{tabular}
\end{table*}

\begin{table*}[h!]
  \caption{Precision (P), recall (R), and F1 score (F1) achieved by models on SponSpeech's test-amb set.}
  \label{tab:results-test-amb}
  \centering
  \begin{tabular}{ c | c c c | c c c | c c c | c c c }
    \toprule
    \multirow{2}{*}{\textbf{Model}} & \multicolumn{3}{c|}{\textbf{Full Stop}} & \multicolumn{3}{c|}{\textbf{Comma}} & \multicolumn{3}{c|}{\textbf{Question}} & \multicolumn{3}{c}{\textbf{Overall}} \\
    & P & R & F1 & P & R & F1 & P & R & F1 & P & R & F1 \\
    \midrule
    EfficientPunct & 78.0 & 82.9 & \textbf{80.4} & 56.5 & 81.6 & 66.7 & 72.1 & 78.8 & \textbf{75.3} & 64.4 & 82.0 & 72.1 \\
    UniPunc & 70.6 & 79.8 & 74.9 & 47.0 & 74.5 & 57.7 & 69.9 & 70.0 & 70.0 & 55.7 & 76.5 & 64.5 \\
    BERT & 81.3 & 79.2 & 80.3 & 65.5 & 73.5 & \textbf{69.3} & 74.6 & 75.5 & 75.0 & 71.9 & 75.9 & \textbf{73.8} \\
    GPT-4 Turbo & 85.5 & 66.8 & 75.0 & 48.8 & 84.3 & 61.8 & 65.8 & 80.0 & 72.2 & 58.4 & 77.0 & 66.4 \\
    \bottomrule
  \end{tabular}
\end{table*}

To initiate punctuation restoration research using SponSpeech, we provide several notable models' baseline results in Table \ref{tab:results-test} for the test set and Table \ref{tab:results-test-amb} for the test-amb set. Training and validation are performed using SponSpeech's standard data subsets. The models picked are:
\begin{itemize}
    \item \textbf{EfficientPunct}, an ensemble with a time-delay neural network to process BERT \cite{devlin2018bert} and Kaldi-derived \cite{povey2011kaldi} embeddings, picked for its state of the art performance and efficiency. Considers text and acoustics \cite{liu2024efficient}.
    \item \textbf{UniPunc}, an attention-based architecture with a coordinate bootstrapper that allows for missing audio. We apply the same BERT and Kaldi embeddings as EfficientPunct for fair comparison of the architectures. Considers text and acoustics \cite{zhu2022unified}.
    \item \textbf{BERT}, a bidirectional transformer-based foundation language model that can be fine-tuned for a wide variety of downstream tasks with very few additional layers. Considers text only \cite{devlin2018bert}.
    \item \textbf{GPT-4 Turbo}, a generative pretrained transformer large language model created by OpenAI, picked due to its immense popularity in the AI community and the public through ChatGPT and its origin model, GPT-4. Considers text only \cite{openai2023gpt}. We evaluate the zero-shot performance of this model.
\end{itemize}

The best performance for all types of punctuation marks in both evaluation sets is dominated by BERT and EfficientPunct.

For the test set, BERT achieves highest F1 scores for full stops, commas, and overall, while EfficientPunct achieves the highest for question marks. BERT's language modeling capabilities for punctuation restoration outmatch all other models. Although, EfficientPunct's outperformance on question marks may be attributed to the consideration of acoustics, which can often signal question marks with rising pitch. BERT may relatively suffer from the lack of pitch/acoustic information in this case.

For the test-amb set, BERT achieves highest F1 scores for commas and overall, while EfficientPunct achieves the highest for full stops and question marks. All models performed worse on the test-amb set than on the test set.

An important finding from the baseline results is that the multimodal models proposed thus far fall short of BERT, which has continually proven to be robust in the task of punctuation restoration. Yet, experiments on MuST-C's highly structured TED talks have repeatedly shown the proposed models' outperformance over BERT. This highlights the importance of testing on diverse datasets.

\section{Conclusion}
\label{sec:conclusion}

This paper introduces SponSpeech, a new dataset for punctuation restoration with spontaneous speech in the form of podcast conversations. Many ASR datasets, especially scripted ones, lack natural tendencies in speech like stutters, random pauses, grammatical imperfections, and interruptions among multiple speakers. We remedy this dearth of important speaking characteristics in datasets by contributing SponSpeech as a public, free resource, licensed under CC BY-NC 4.0.

Finally, we discuss some limitations and directions for future research. It should be noted that certain punctuation decisions are stylistic, and oftentimes two or more ways of punctuating are readable, reasonable, and grammatically correct, all without altering the text's intended meaning. Such subjectivities slightly influence punctuation labels, and, in turn, evaluation results. Measures of punctuation accuracy are hence coarser than one may initially assume. Also affected are our labeling of the training and validation set for ELECTRA fine-tuning. Especially in cases of spontaneous speech when grammatical rules are defied, flawless and objective punctuation labeling proves even more difficult. Our classification of whether select utterances are ambiguous may somewhat suffer from this linguistic phenomenon.

As for potential future work, multimodal models falling short of BERT on SponSpeech serves as strong motivation for further research in multimodal punctuation restoration models. It has been repeatedly shown that multimodal approaches have significant advantages over text-only models \cite{yi2019self}. We can therefore conclude that there exists effective, undiscovered methods of considering acoustics for punctuation restoration. Most likely candidates involve heavily weighting a language model's grammatical insight, while leveraging acoustics for fine-grained information.

\bibliographystyle{IEEEbib}
\bibliography{ms}

\end{document}